# An Adaptive Quantum-inspired Differential Evolution Algorithm for 0-1 Knapsack Problem

Ashish Ranjan Hota
Department of Electrical Engineering
Indian Institute of Technology
Kharagpur, India
hota.ashish@acm.org

Ankit Pat
Department of Mathematics
Indian Institute of Technology
Kharagpur, India
ankitpat.iitkgp@gmail.com

*Abstract* — Differential evolution (DE) is a population based evolutionary algorithm widely used for solving multidimensional global optimization problems over continuous spaces. However, the design of its operators makes it unsuitable for many real-life constrained combinatorial optimization problems which operate on binary space. On the other hand, the quantum inspired evolutionary algorithm (QEA) is very well suitable for handling such problems by applying several quantum computing techniques such as Q-bit representation and rotation gate operator, etc. This paper extends the concept of differential operators with adaptive parameter control to the quantum paradigm and proposes the adaptive quantum-inspired differential evolution algorithm (AQDE). The performance of AQDE is found to be significantly superior as compared to QEA and a discrete version of DE on the standard 0-1 knapsack problem for all the considered test cases.

## I. INTRODUCTION

Differential Algorithm (DE), introduced by Storn and Price [1,2] has been shown to give significantly better performance in terms of efficiency and robustness on many benchmark multimodal continuous functions than other population based evolutionary algorithms. For exploration of the search space and to introduce diversity, it employs two simple mutation and crossover operators respectively followed by a greedy replacement strategy. The performance is found to be very sensitive to the mutation and crossover parameters chosen and the best combination of both the parameters changes from one function to another. Thus, a large number of modifications have been proposed to make the selection of control parameters adaptive and free from function dependency [3-7].

Because of its superior performance on continuous optimization problems, several modifications have been introduced in the past, so that it operates on binary space. Pampara, Engelbrecht and Franken [8] proposed an angle modulation scheme (AMDE) to map the continuous space to binary. On similar lines, binary differential evolution (binDE) and normalization DE (normDE) were proposed based on sigmoid function mapping and normalization of continuous space respectively [9] giving better results as compared to AMDE. A discrete binary version of differential evolution (DBDE) for solving 0-1 knapsack problem was also proposed [10].

To solve various optimization problems better than the conventional evolutionary algorithms, a broad class of algorithms have been proposed by applying several concepts of quantum computing in the past decade. Quantum computing uses the quantum mechanical phenomena like superposition, entanglement, interference, de-coherence, etc to develop quantum algorithms. Many quantum algorithms have been shown to be exponentially faster and massively parallel as compared to classical algorithms [11, 12]. Thus quantum inspired genetic algorithms with interference as crossover operator [13], quantum inspired evolutionary algorithms (QEA) [14], quantum behaved particle swarm optimization [15] etc has been developed for both continuous and binary spaces.

QEA uses superposition of binary bits known as Q-bit for representation of individuals and updates the individuals depending on their values with respect to the global best solution by suitably deciding the parameter of the rotation gate operator. Broadly it comes under the class of estimation of distribution algorithms (EDA) [23]. QEA has demonstrated quite significant results on binary optimization problems and some improvements on QEA have also been proposed [16, 17]. QEAs have been extended by differential operators to solve flow shop scheduling problems [18], N-queen's problem [19], for classification rule discovery [20] and some benchmark functions [21].

In this paper, an adaptive quantum-inspired differential evolution algorithm (AQDE) is proposed with adaptive control of mutation and crossover parameters and the operators acting directly on the superposition states of the individual. The proposed AQDE outperforms QEA and DBDE under different conditions of population size and item size of the 0-1 knapsack problem.

The rest of this paper is organized as follows: Section II gives a brief introduction of knapsack problem, DE, DBDE and QEA. The proposed AQDE is explained in detail in section III. Experimental settings and the results obtained are mentioned under section IV. Finally, section V concludes the paper.

## II. BACKGROUND

### A. 0 - 1 Knapsack Problem:

The 0-1 knapsack problem is a classical problem in combinatorial optimization.
Problem Description:

In a given set of m items each item has an integer weight $w_j$ and an integer profit $p_j$. The problem is to select a subset from the set of m items such that the overall profit is maximized without exceeding a given weight capacity W. It is an NP-Hard problem and hence doesn't have a polynomial time algorithm. The problem may be mathematically modeled as follows:

Maximize: $\sum_{i=1}^{m} p_i x_i$ (1)

Subject to the constraint: $\sum_{i=1}^{m} w_i x_i \leq W, x_i \in \{0,1\}$ (2)

where $x_i$ takes values of either 1 or 0 representing the selection or rejection of the $i^{th}$ item.

### B. Differential Evolution :

In classical DE, each member of the population is represented by a real valued D-dimensional vector. A typical iteration of the DE algorithm consists of three major operations – mutation, crossover and selection, which are carried out for each member of the population (called as target vector). Mutation on each target vector of the population generates a new mutant vector uniquely associated with it. Then the crossover operation generates a new trial vector using the mutant vector and the target vector itself. In selection phase the fitness of the trial vector is compared with the target vector and the vector with higher fitness replaces the target vector in the population for the next iteration. The three operations – mutation, crossover and selection, are discussed in detail below.

Mutation: The mutant $V_i^t$ vector on a target vector $X_i^t$ is generated by adding a randomly selected vector $X_{r1}^t$ from the population, with a weighted difference of two other randomly selected vectors $X_{r2}^t, X_{r3}^t$ from the population.

$$V_i^t = X_{r1}^t + F.(X_{r2}^t - X_{r3}^t)$$ (3)

where r1,r2 and r3 are all distinct and different from i. The parameter t denotes the generation. F is a control parameter whose value is typically chosen between 0 and 2.

Crossover: The crossover operation generates a trial vector $U_i$ from its corresponding target vector $X_i$ and mutant vector $V_i$, by using the following relation:

$$u_{j,i}^t = \begin{cases} v_{j,i}^t, & if\ (rand_j(0,1) \leq CR)\ or\ (j = I_{rand}) \\ x_{j,i}^t, & if\ (rand_j(0,1) > CR)\ and\ (j \neq I_{rand}) \end{cases}$$ (4)

where j=1,2,…..D, $U_i = (u_{1,i}^t, u_{2,i}^t, ……., u_{D,i}^t)$, $rand_j$ is the jth evaluation of a random number generator in [0,1] from a uniform distribution. $I_{rand}$ is a randomly chosen dimension index from {1,2,…..,D} which ensures that the new trail vector is different from the target vector. CR is a control parameter which decides the crossover rate and its value is typically chosen in the range of 0 to 1.

Selection: If the trial vector $U_i$ has a better fitness value compared to the target vector, then it replaces the target vector in the population in the next iteration. Otherwise, the target vector remains unchanged in the population.

### C. Discrete Binary version of Differential evolution

The discrete binary version of differential evolution (DBDE) [10] was an attempt to develop an algorithm which worked on similar lines as DE but on a binary D-dimensional space. DBDE has its roots in DE and a discrete binary version of particle swarm optimization (DPSO) [22].

Here the individuals are initialized as a binary string. The mutation operator is exactly similar to that of DE, but the resultant mutant vector is no longer binary because of the difference operator and the control parameter. Therefore the discretization process from a real continuous space to a binary space is done according to the following equation:

$$v_{i,d}^t = \begin{cases} 1, & if\ rand(0,1) \leq sig(v_{i,d}^t) \\ 0, & if\ rand(0,1) > sig(v_{i,d}^t) \end{cases}$$ (5)

where *rand* is a random number in the range [0,1] selected uniformly at random. sig() is a sigmoid limiting transformation function and $v_{i,d}^t$ is $d^{th}$ dimensional value of the $i^{th}$ mutated vector in generation t. The crossover and selection operations in DBDE, are same as in DE.

### D. Quantum- Inspired Evolutionary Algorithm

Quantum-inspired Evolutionary Algorithm (QEA), as its name indicates, is inspired from the principles of quantum computing, but it is designed to run on a classical computer.

In QEA, the smallest unit of information is called Q-bit and is defined as $[\alpha,\beta]^T$, where α and β are complex numbers that specify the probability amplitude of the respective Q-bit states such that $|\alpha|^2+|\beta|^2 =1$. $|\alpha|^2$ represents the probability that the Q-bit will be in state '0' and $|\beta|^2$ represents the probability that the Q-bit will be in state '1'. The representation for an individual q of QEA with m-bit is given as follows:

$$q = \begin{bmatrix} \alpha_1 & \alpha_2 & \cdots & \alpha_m \\ \beta_1 & \beta_2 & \cdots & \beta_m \end{bmatrix}$$ (6)

where $|\alpha_i|^2+|\beta_i|^2 =1$, i=1,2,…..m

Algorithm Description: In the beginning, the population is initialized with the α and β of all bits of all individuals set to $1/\sqrt{2}$. In each generation, binary strings are generated from the respective Q-bit strings by observing the Q-bit states using the following criteria:

$$P_{i,j} = \begin{cases} 1, & if\ rand() < |\beta_{i,j}|^2 \\ 0, & otherwise \end{cases}$$ (7)

where $P_{i,j}$ is the jth bit of $i^{th}$ individual in the population. Once the population consisting of the binary strings has been generated, the fitness value of these strings is evaluated and the best solutions are stored separately in a global pool **B**.

The global best solution **b** among all the solutions in **B** is determined. Then, a quantum rotation gate U(θ) is used to update the values of the Q-bits of each individual as follows:

$$U(\Delta\theta_i) = \begin{pmatrix} \cos(\Delta\theta_i) & -\sin(\Delta\theta_i) \\ \sin(\Delta\theta_i) & \cos(\Delta\theta_i) \end{pmatrix} \quad (8)$$

where $\Delta\theta_i$, i=1,2,....m is the rotation angle of each Q-bit towards either 0 or 1 depending on its sign. The parameter $\Delta\theta_i$ is decided by comparing the value of the bit in the individual and the corresponding bit in the global best individual as per Table I (reproduced from [14]).

Table I. Look up Table for $\Delta\theta_i$
(f(.) is the profit and $b_i$ and $x_i$ are $i^{th}$ bit of best solution **b** and binary solution **x**)

| $x_i$ | $b_i$ | $f(x) \geq f(b)$ | $\Delta\theta_i$ |
|---|---|---|---|
| 0 | 0 | false | 0 |
| 0 | 0 | true | 0 |
| 0 | 1 | false | $0.01\pi$ |
| 0 | 1 | true | 0 |
| 1 | 0 | false | $-0.01\pi$ |
| 1 | 0 | true | 0 |
| 1 | 1 | false | 0 |
| 1 | 1 | true | 0 |

Then the global pool is updated with fitness based replacement by better individuals of the present generation and the previous global pool. The global best individual is also updated accordingly. A global and local migration is invoked with a definite frequency, in which all or some of the individuals of the global pool are replaced by the global best or the local best individuals respectively.

The detailed procedure of QEA [14] is provided below for better understanding.

### III. ADAPTIVE QUANTUM DIFFERENTIAL EVOLUTION

This section describes the adaptive quantum-inspired differential evolution algorithm (AQDE).

#### A. Representation

Instead of using $[\alpha,\beta]^T$ like QEA as the representation of Q-bits, AQDE uses the variable θ for reasons discussed later. Since $|\alpha|^2 + |\beta|^2 = 1$, it basically represents the equation of a unit circle and each point on its perimeter can be represented by a single variable θ with the Cartesian co-ordinates given by cosθ and sinθ where θ is defined in [0,2π]. In AQDE, the Q-bits (θ) are initialized uniformly at random in [0, 2π] for all the bits for all the individuals in the population. The binary population is derived as follows:

$$P_{j,i} = \begin{cases} 1, & if \quad rand(0,1) < \sin^2(\theta_{j,i}) \\ 0, & otherwise \end{cases} \quad (9)$$

where $P_{j,i}$ is the $j^{th}$ bit of the $i^{th}$ individual in the population and $\theta_{j,i}$ is the corresponding Q-bit.

```
Procedure QEA
begin
  t ← 0
  initialize Q(t);
  make P(t) from Q(t) by (7)
  evaluate P(t)
  B(t) ← P(t)
  b← best solution among B(t)
  while t<T do
     t←t+1
     make P(t) from Q(t) by (7)
     evaluate P(t)
     update Q(t) using (8)
     store best solutions among B(t-1)
     and P(t)in B(t)
     store best solution b among B(t)
     if (migration condition)
        migrate b or bᵗⱼ to B(t) globally
        or locally respectively
     endif
  end while
end
```

Figure 1. QEA pseudo code

#### B. Mutation Operator

Mutation operator in AQDE is similar to that of classical DE, but instead of operating on the individual directly, it is applied on the Q-bit (θ). Since θ contains information about both α and β, it is more appropriate to generate the mutant vector in terms of θ. Moreover, unlike the case of classical DE, it inherently avoids the problem of constraint violation, i.e. the mutant vector exceeding the prescribed domain. This is because both cosine and sine functions are periodic with period 2π. The representation of Q-bits was changed to θ keeping this in mind. The mutant Q-bits $\theta^m$ are generated for all the individuals in the population in every generation. Mutant Q-bits of the $i^{th}$ individual in generation t are determined as follows:

$$\theta_i^{mt} = \theta_{r1}^t + F^t.(\theta_{r2}^t - \theta_{r3}^t) \quad (10)$$

where r1,r2,r3 and i are mutually distinct and $F^t$ is the mutation control parameter which is determined in every generation as per the following equation,

$$F = rand_1.rand_2.(0.1) \quad (11)$$

where $rand_1$, $rand_2$ are random numbers generated from a uniform distribution on [0,1]. The purpose of multiplying one random number is to take values for F on the interval [0,0.1]. One more independent random number is further multiplied to probabilistically generate more values close to zero. This is because, the quality of solution is found to be highly sensitive towards radical perturbation of the Q-bit.

#### C. Crossover Operator

The crossover operation operates on the original Q-bits and the respective mutant Q-bits in the following

manner:

$$\theta_{j,i}^{ct} = \begin{cases} \theta_{j,i}^{mt}, & if(rand_j(0,1) \leq CR^t) or(j = I_{rand}) \\ \theta_{j,i}^{t}, & if(rand_j(0,1) > CR^t) and(j \neq I_{rand}) \end{cases} \quad (12)$$

where $\theta_{ji}^c$ is the jth Q-bit of ith individuals after the crossover operation. $I_{rand}$ is a number randomly chosen from {1,2,…D} which ensures at least one Q-bit is different from the original set in each individual. $CR^t$ is the control parameter which is determined in every iteration as follows:

$$CR^t = G_{rand}(0.5, 0.0375) \quad (13)$$

where $G_{rand}$ generates a random number from the Gaussian distribution with mean 0.5 and standard deviation 0.0375. As a result, CR lies in a 0.15 neighborhood of 0.5 with a probability of 0.998. Thus the value of CR is selected very close to 0.5 in almost all the cases, which is found to the best value experimentally.

*D. Selection*

The population and the Q-bits are updated in a greedy fashion. By observing the state of the newly obtained Q-bits ($\theta_{ji}^c$), by (9), a new set of individuals are obtained which replace the corresponding individual in the population if their fitness values are higher. The replacement is done using the following equations:

$$P_i^{t+1} = \begin{cases} P_i^{ct}, & if(f(P_i^{ct}) > f(P_i^t)) \\ P_i^t, & otherwise \end{cases} \quad (14a)$$

and

$$\theta_{j,i}^{t+1} = \begin{cases} \theta_{j,i}^{ct}, & if(f(P_i^{ct}) > f(P_i^t)) \\ \theta_{j,i}^{t}, & otherwise \end{cases} \quad (14b)$$

where $P_i^c$ is the ith individual by observing the Q-bits modified after crossover ($\theta_{ji}^c$). $f(P_i)$ is the fitness value of the corresponding individual.

```
Procedure AQDE for knapsack
begin
  t ← 0
  initialize Q(t);
  make P(t) from Q(t) by (9)
  repair P(t)
  evaluate fitness of P(t)
  while t<T  do
     t←t+1
     determine F and CR by (11) and (13)
     apply mutation on Q(t) using (10)
     obtain Q'(t) by crossover using (12)
     make P'(t) from Q'(t) using (9)
     repair P'(t)
     evaluate fitness of P(t)
     update P(t+1) and Q(t+1) by (14)
  end while
end
```
Figure 2. AQDE pseudo code

Thus AQDE is an adaptive algorithm, which employs differential operators on the superposition state of Q-bits and can be applied to binary optimization problems directly. The pseudo code of AQDE applied to 0-1 knapsack problem is given in Fig. 2.

IV. EXPERIMENTAL SETTINGS AND RESULTS

To test the performance of AQDE, it was compared with both QEA and DBDE on the 0-1 knapsack problem. In all test cases, strongly correlated sets of data were considered. The weights $w_i$, respective prices $p_i$ and the knapsack capacity W were calculated as follows [14].

$$w_i = rand[1,10]$$
$$p_i = w_i + 5, \quad i = 1,2,.....m \quad (15)$$
$$W = \frac{1}{2}\sum_{i=1}^{m} w_i$$

where rand[1,10] generates an integer in {1,2,….,10} uniformly at random.

For satisfying the constraint of the knapsack problem, the repair method given in [14] is applied to all the algorithms. If the constraint is violated, the repair method randomly chooses an item and removes it from the collection until the constraint is just satisfied. After that it starts adding items randomly again. When the constraint is just violated, it removes the last added item and stops.

Three knapsack problems with 100, 250, and 500 items were considered with unsorted data obtained as above. For each knapsack problem, the algorithm was tested for a population size of 30 and 50. The maximum number of generations in all cases was chosen as 1000. The mean best profits of 30 runs and the respective standard deviations were tabulated (Table I).The variation of mean best profit with no. of generations were plotted (Fig. 3-8).

Table II. Performance comparison on 0-1 Knapsack problem

| Item size | QEA | | DBDE | | AQDE | |
|---|---|---|---|---|---|---|
| | 30 | 50 | 30 | 50 | 30 | 50 |
| 100 | 600.5 (4.615) | 601.8 (3.333) | 612.9 (3.208) | 614.0 (2.936) | 625.6 (3.301) | 629.1 (3.147) |
| 250 | 1423.9 (8.758) | 1428.3 (5.427) | 1448.1 (5.708) | 1452.5 (6.301) | 1502.7 (6.834) | 1519.4 (5.469) |
| 500 | 2693.8 (10.77) | 2703.3 (3.282) | 2735.2 (10.54) | 2740.6 (8.204) | 2855.2 (14.93) | 2898.7 (11.27) |

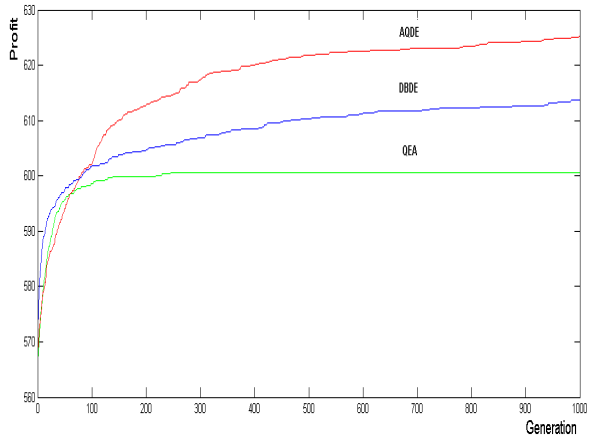

Figure 3. Population Size= 30, Item Size = 100

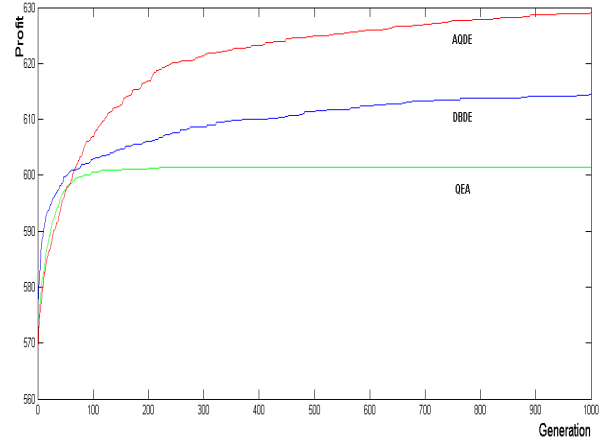

Figure 4. Population Size= 50, Item Size = 100

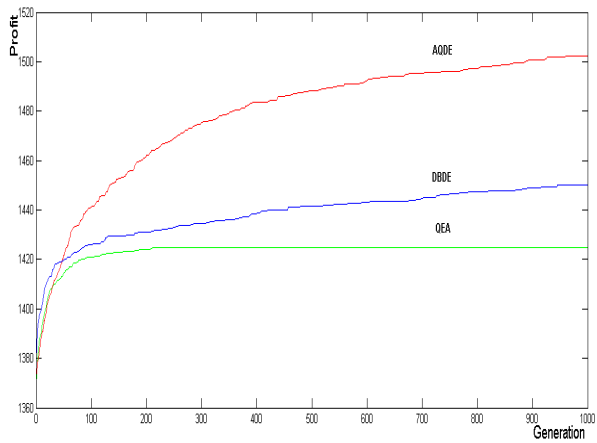

Figure 5. Population Size= 30, Item Size = 250

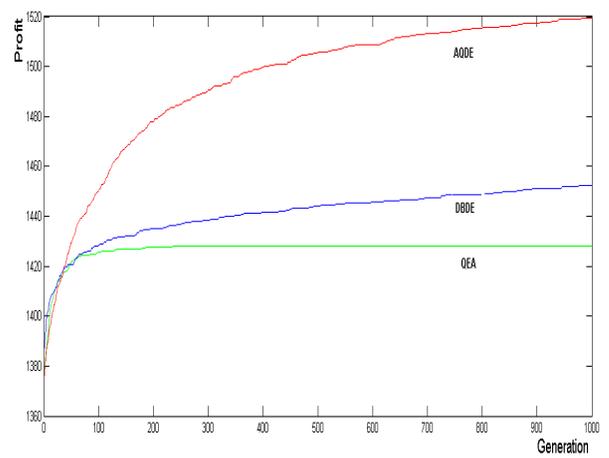

Figure 6. Population Size= 50, Item Size = 250

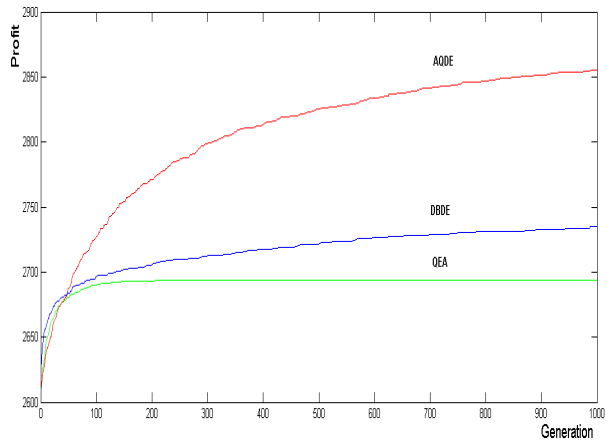

Figure 7. Population Size= 30, Item Size = 500

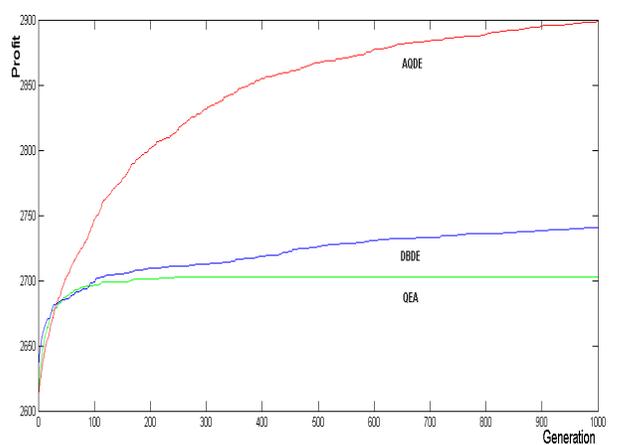

Figure 8. Population Size= 50, Item Size = 500

Figure 3-8 show the progress of the convergence by depicting the average of best profits over 30 runs for the previously mentioned population sizes and item sizes. For all the cases considered, the curve of mean best profit for AQDE lies slightly below the curves of QEA and DBDE for the initial 50 generations, but soon after that, it goes above the curves of QEA and DBDE, thereby showing significantly better results. The plots suggest a premature convergence of both QEA and DBDE as compared to AQDE.

## V. CONCLUSION

In this paper, we have proposed a novel AQDE algorithm for solving the 0-1 Knapsack problem. The proposed algorithm is a hybrid of QEA and DE along with a novel adaptive parameter control method. The experimental results have proved the superior performance of AQDE compared to QEA and DBDE. Here, the performance of AQDE was tested only on the 0-1 Knapsack problem. With some modifications, the concept of the algorithm may be extended to other discrete combinatorial optimization problems.